\documentclass{article}

\usepackage{PRIMEarxiv}

\usepackage[utf8]{inputenc} 
\usepackage[T1]{fontenc}    
\usepackage{hyperref}       
\usepackage{url}            
\usepackage{booktabs}       
\usepackage{amsfonts}       
\usepackage{nicefrac}       
\usepackage{microtype}      
\usepackage{lipsum}
\usepackage{fancyhdr}       
\usepackage{graphicx}       
\usepackage{amsmath}
\usepackage{amssymb}
\graphicspath{{media/}}     

\title{Hybrid ConvNeXt--Swin Model for AI Image Detection}

\author{
  Raleesa Perera \\
  Department of Computer Science and Engineering \\
  University of Moratuwa \\
  Moratuwa, Sri Lanka \\
  \texttt{raleesa.22@cse.mrt.ac.lk}
  \And
  Subhagya Dehiwala Liyanage \\
  Department of Computer Science and Engineering \\
  University of Moratuwa \\
  Moratuwa, Sri Lanka \\
  \texttt{subhagya.22@cse.mrt.ac.lk}
  \And
  Madhuni Karunaratne \\
  Department of Computer Science and Engineering \\
  University of Moratuwa \\
  Moratuwa, Sri Lanka \\
  \texttt{madhuni.22@cse.mrt.ac.lk}
  \And
  Tharani Jayaratne \\
  Department of Computer Science and Engineering \\
  University of Moratuwa \\
  Moratuwa, Sri Lanka \\
  \texttt{tharani.22@cse.mrt.ac.lk}
  \And
  Niduni Kasige \\
  Department of Computer Science and Engineering \\
  University of Moratuwa \\
  Moratuwa, Sri Lanka \\
  \texttt{niduni.22@cse.mrt.ac.lk}
}

\begin{document}
\maketitle

\begin{abstract}
The rapid advancement of generative artificial intelligence has made it possible to produce highly realistic images that can be difficult to distinguish from authentic ones. This study presents a deep learning framework for detecting AI-generated images by combining complementary visual representations. To improve the robustness of the model, data augmentation is applied during training, while a four-channel input representation is introduced by combining the original RGB channels with an additional grayscale channel. This allows the model to capture both color information and luminance-based structural patterns that may help reveal subtle image artifacts. The proposed framework employs two pretrained backbones, ConvNeXt-Tiny and Swin-Tiny, in parallel to learn complementary features, with ConvNeXt-Tiny focusing on local spatial patterns and Swin-Tiny capturing broader contextual relationships. An attention-based feature fusion module is then used to adaptively combine the features learned by the two networks before classification. The fused representation is finally passed to a fully connected classifier to distinguish between AI-generated and human-created images. Experimental results show that the proposed approach can identify subtle visual artifacts and structural inconsistencies while achieving better generalization than the individual backbone models. Overall, the findings demonstrate the potential of combining multi-channel image representations with complementary CNN and Transformer-based features for more reliable AI-generated image detection.
\end{abstract}

\keywords{AI-generated image detection, Deep learning, ConvNeXt, Swin Transformer, Feature fusion, Data augmentation}

\section{Introduction}
Recent advancements in generative artificial intelligence have enabled the creation of highly realistic synthetic images that are increasingly difficult to distinguish from authentic images. This has created some concerns about misinformation, digital forensics, and image content verification~\cite{keita2024fidavl}.

As generative methods become more advanced, the artifacts and inconsistencies in images generated using AI might be very subtle. Existing methods may also struggle to generalize to images produced by unseen generation techniques~\cite{cao2024hyperdet}. These challenges highlight the need for detection methods that can capture diverse visual characteristics and generalize across different image sources.

To address these challenges, this study proposes a deep learning framework that combines a four-channel RGB+Grayscale input representation with complementary feature extraction using ConvNeXt-Tiny and Swin-Tiny. The additional grayscale channel provides luminance and structural information, while the two backbones capture complementary local and global image features. An attention-based fusion mechanism combines these representations before classification.

The main contributions of this work are:
\begin{itemize}
\item A four-channel RGB+Grayscale input representation for enhanced visual feature representation.
\item A hybrid ConvNeXt-Tiny and Swin-Tiny architecture for complementary feature extraction.
\item An attention-based feature fusion mechanism for combining the learned representations.
\item Ablation and cross-dataset evaluations to assess model performance and generalization.
\end{itemize}

\section{Related Work}

Early research in the field established that standard deep learning architectures could be surprisingly effective at identifying synthetic content. A landmark study by Wang et al.~\cite{wang2020cnn} demonstrated that a standard ResNet-50 model, when trained with extensive data augmentation, could generalize across various generative models to detect synthetic artifacts. This work proved that Convolutional Neural Networks (CNNs) were capable of learning common flaws shared by many generators, serving as a primary baseline for modern detection research. Building on this, Gragnaniello et al.~\cite{gragnaniello2021gan} emphasized that the choice of architecture and the inclusion of frequency domain features were critical for robust performance. Their findings highlighted how texture-based artifacts, often better captured in the frequency domain or via grayscale representations, provided vital clues that standard color-based analysis might overlook. By prioritizing these structural details, their work justified the shift toward multi-component input representations to improve detection accuracy.

The evolution of AI image detection saw a shift from purely CNNs to hybrid architectures that leveraged the complementary strengths of local and global feature extraction. Initially, ConvNeXt, introduced by Liu et al.~\cite{liu2022convnet}, demonstrated that modernized CNN architectures could match or exceed the performance of Vision Transformers (ViTs) by optimizing standard ResNet blocks with larger kernels and inverted bottlenecks.EfficientNet~\cite{tan2020efficientnet} introduced a systematic compound scaling method that uniformly balances network depth, width, and resolution, enabling state-of-the-art accuracy with significantly lower computational costs than traditional CNNs. Conversely, DINOv2~\cite{oquab2024dinov2} utilizes Vision Transformers trained via self-supervised distillation on massive curated datasets to produce robust, task-agnostic representations. In the context of AI-generated image detection, ConvNeXt served as a robust backbone for capturing fine-grained structural artifacts and local pixel inconsistencies characteristic of generative model upsampling. Parallelly, the Swin Transformer~\cite{liu2021swin} introduced a hierarchical representation using shifted windowing mechanisms. Unlike global Transformers, Swin Transformers were computationally efficient and excelled at capturing multi-scale global context. This was particularly vital for identifying semantic inconsistencies in synthetic images that appeared locally correct but globally incoherent.

Recent advancements focused on the effective combination of these two paradigms. Studies such as ConvTransGFusion~\cite{ruan2026transformer} demonstrated that bridging the inductive biases of CNNs with the long-range dependency capabilities of Transformers through attention-based gating mechanisms significantly enhanced robustness. Such hybrid frameworks provided superior performance in image forensics by simultaneously analyzing local texture anomalies (via ConvNeXt) and global structural logic (via Swin Transformer), establishing a new state-of-the-art for detecting high-quality AI-generated content.

While existing methods were effective, they often treated color and texture analysis as separate, sequential steps, frequently converting images to different formats before processing. This fragmented approach could lead to a loss of vital information where color and structural inconsistencies overlapped. In contrast, our proposed 4-channel framework offered a more integrated alternative. By processing RGB and Grayscale data simultaneously from the start, the model preserved both rich color information and fine-grained structural artifacts throughout the entire pipeline. This unified input ensured that the attention-based fusion mechanism could directly learn the complex relationships between color shifts and texture anomalies. Ultimately, this approach moved beyond isolated feature extraction, providing a more comprehensive and robust strategy for identifying the subtle flaws in high-quality, modern AI-generated imagery. Table I emphasizes how the above mentioned models perform when it comes to modern AI-generated image detection.

\begin{table}[t]
  \caption{Comparison of F1-Scores of the Models Studied}
  \centering
  \begin{tabular}{lcc}
    \toprule
    Model & Year & F1-score \\
    \midrule
    ResNet 50 & 2015 & 0.3047 \\
    DenseNet & 2017 & 0.6558 \\
    MaxVit & 2022 & 0.6534 \\
    RegNet & 2020 & 0.3056 \\
    DINOv2 & 2023 & 0.7352 \\
    EfficientNet & 2019 & 0.7451 \\
    ConvNeXt + Swin Transformer & 2026 & 0.7681 \\
    \bottomrule
  \end{tabular}
  \label{tab:classification_report}
\end{table}

\section{Dataset}

For this research, the AI vs. Human-Generated Images dataset~\cite{ai_vs_human_dataset} from Kaggle was selected. This dataset was comprised of images obtained from the Shutterstock website and belonged to various categories. To ensure diversity, about one-third of the images included human subjects. Each real image was paired with a corresponding AI-generated version created using modern generative models.

The training dataset contained 79,950 images, while the test dataset contained 5,540 images. The original training data was further split into training and validation sets, where 20\% of the data was used for validation. Stratified sampling was applied to maintain the same label distribution in both sets.

\section{Methodology}
\subsection{Overview}

The proposed system was designed to distinguish between AI-generated and human-created images. First, the input images were augmented to increase diversity and improve model robustness. They were then converted to a tensor and extended into a 4-channel representation (RGB + grayscale) to capture both color and structural details. The 4-channel tensors were fed into two parallel feature extraction networks, ConvNeXt and Swin Transformer, which learned complementary features. These features were fused using an attention module to focus on the most important information, and the combined features were finally passed through a fully connected classifier to make the prediction.

\begin{figure}
  \centering
  \includegraphics[width=\textwidth]{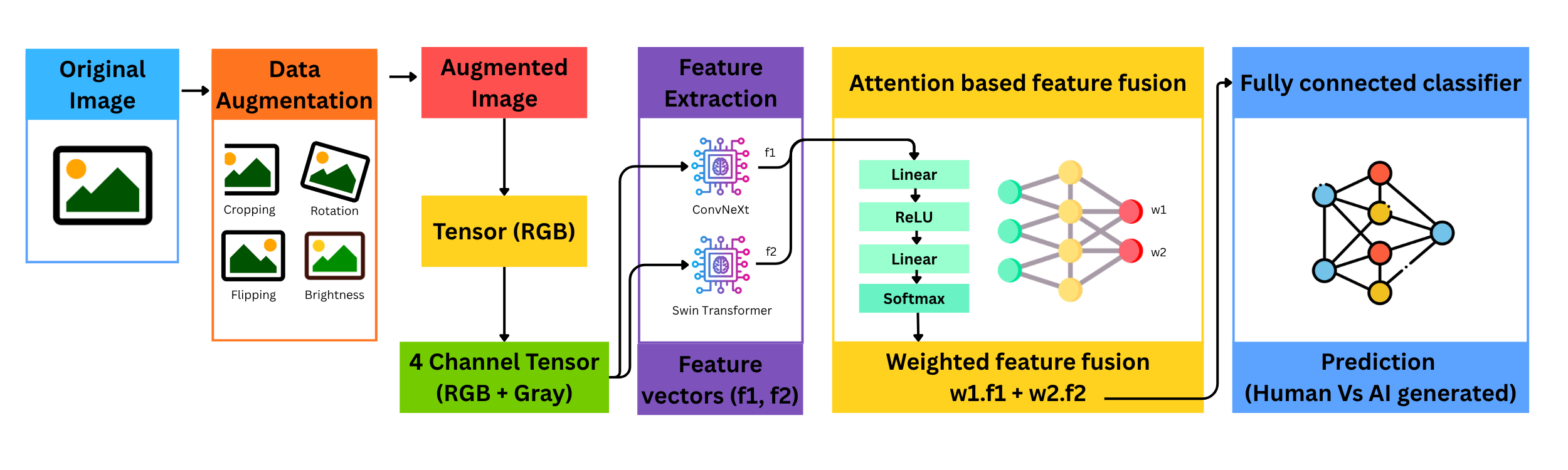}
  \caption{Architecture of the Proposed AI vs Human Image Classification Model.}
  \label{fig:fullwidth}
\end{figure}

\subsection{Data Augmentation}

To improve the generalization capability and robustness of the proposed model, several data augmentation techniques were applied to the training images prior to model training. The applied techniques included Random Resized Crop, Horizontal Flip, Rotation, Color Jitter, Random Perspective, Gaussian Blur, Gaussian Noise Injection, and Autocontrast. By exposing the model to variations in image orientation, scale, lighting conditions, and visual quality, these augmentations helped prevent overfitting and enabled the model to better generalize to previously unseen AI-generated patterns and real-world images.

\begin{figure}
  \centering
  \includegraphics[width=\textwidth]{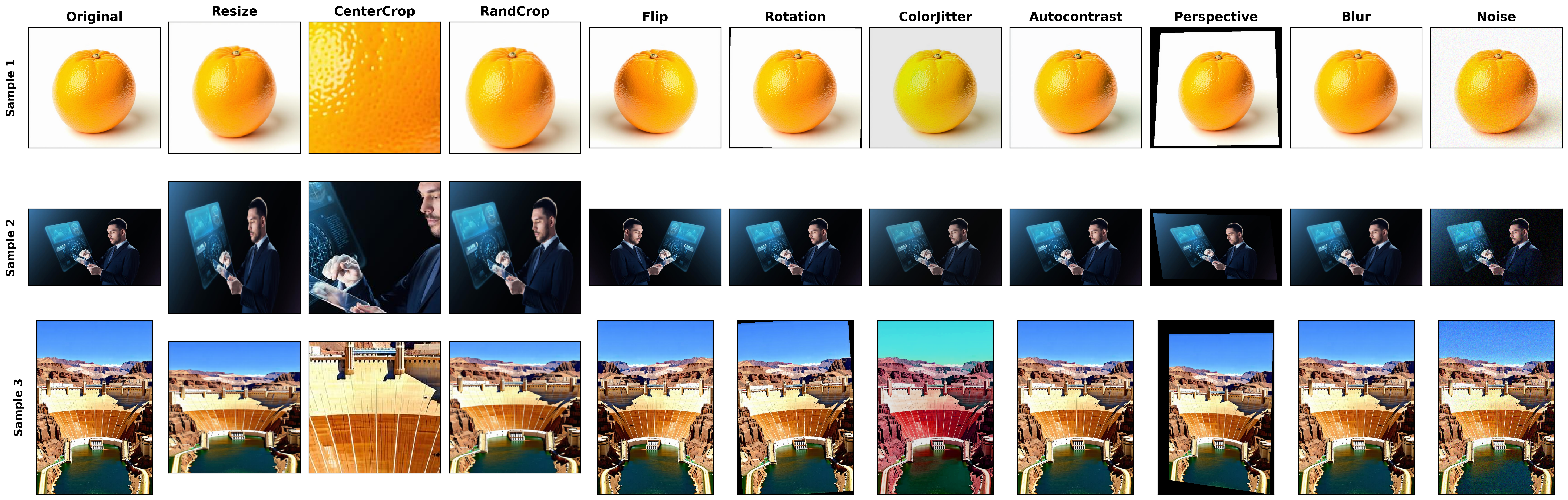}
  \caption{Data augmentation applied to each image. The first three rows show augmented versions of human-created images, while the next three rows show augmented versions of AI-generated images. Each row demonstrates the variety of transformations applied to the original images, such as rotation, flipping, and color adjustments.}
  \label{fig:augmentation}
\end{figure}

\subsection{Input Representation using RGB + Grayscale}

To enhance feature representation, the conventional three-channel RGB input was extended to a four-channel format by incorporating an additional grayscale channel. The grayscale channel was computed as the mean intensity of the RGB channels and concatenated with the original image, resulting in a unified four channel input tensor.

Both ConvNeXt-Tiny and Swin-Tiny were pretrained on ImageNet using three-channel RGB images. To allow them to process the proposed RGB+Grayscale representation, the input layer of each backbone was modified from three input channels to four input channels while maintaining the same 96 output channels. The additional grayscale channel was generated by calculating the mean of the R, G, and B values at each pixel and then concatenating this grayscale image with the original RGB channels. Thus, each input image was represented as a 4-channel tensor containing red, green, blue, and grayscale information. The pretrained weights for the original three RGB channels were retained without modification. Since no pretrained weights were available for the additional grayscale channel, its weights were initialized using the mean of the pretrained weights across the three RGB channels. This provided a meaningful initialization for the additional channel rather than using random weights, while preserving the knowledge learned by both backbones during ImageNet pretraining.

By integrating those information, the proposed four-channel representation provided complementary features that improved the model's ability to distinguish between AI-generated and authentic images. 

\subsection{Feature Extraction}

To extract meaningful representations from the input images, two different backbone networks were used: ConvNeXt-Tiny and Swin-Tiny. ConvNeXt-Tiny was a convolutional neural network (CNN) that was effective at capturing local spatial patterns such as textures and edges. In contrast, Swin-Tiny was a transformer-based vision model that could capture global contextual relationships within the image through self-attention mechanisms. By using both architectures in parallel, the model was able to learn both local visual features and long-range dependencies, which improved its ability to distinguish between AI-generated and human-created images.

The individual and combined contributions of these two backbone architectures were systematically evaluated in the ablation study in Section ~\ref{sec:ablation}

\subsection{Feature Fusion}

To effectively combine complementary representations learned by the two backbone networks, an attention-based feature fusion mechanism was employed. Let $f_1 \in \mathbb{R}^{768}$ and $f_2 \in \mathbb{R}^{768}$ denote the feature vectors extracted from ConvNeXt-Tiny and Swin-Tiny, respectively. These features captured different aspects of the input image due to the architectural differences between convolutional and transformer-based models.

The two feature vectors were first concatenated to form a joint representation:
\begin{equation}
f = [f_1; f_2] \in \mathbb{R}^{1536}
\end{equation}

The combined feature vector was then passed through a lightweight attention network consisting of two fully connected layers with a ReLU activation:
\begin{equation}
z = W_2 \, \sigma(W_1 f)
\end{equation}

\begin{equation}
w_i = \frac{\exp(z_i)}{\sum_{j=1}^{2} \exp(z_j)}
\end{equation}

where $w = [w_1, w_2] \in \mathbb{R}^{2}$ represents the normalized attention weights corresponding to each feature stream, $\sigma$ denotes the ReLU activation function, and $W_1$ and $W_2$ are learnable parameters.

The final fused feature representation is computed as a weighted sum of the individual features:

\begin{equation}
f_{\text{fused}} = w_1 f_1 + w_2 f_2
\end{equation}

This attention-based fusion strategy enables the model to adaptively emphasize the more informative feature representation for each input sample. Consequently, the fused representation captures both local texture patterns (from ConvNeXt) and global contextual relationships (from Swin Transformer), improving the distinction between AI-generated and human-created images.

\subsection{Classification and Prediction}

The fused feature representation was passed to a fully connected neural network that performed the final classification. The classifier consisted of linear layers followed by activation functions and dropout for regularization. During training, the CrossEntropyLoss function was used to measure the difference between the predicted labels and the true labels. The model parameters were optimized using the Adam optimizer, which efficiently updated the network weights to minimize the training loss. The model was trained with the following hyperparameters.
\begin{itemize}
    \item Epochs: 2
    \item Normalization mean: [0.485, 0.456, 0.406, 0.45]
    \item Normalization std: [0.229, 0.224, 0.225, 0.225]
    \item Feature dimension (per backbone): 768
    \item Dropout rate: 0.3
    \item Learning rate: le-4
\end{itemize}

During inference, the trained model output the probability of each input image belonging to one of two classes. The final prediction was obtained by selecting the class with the highest probability.

\section{Results}

\subsection{Evaluation Metrics}
To evaluate the performance of the proposed model, accuracy, precision, recall, F1-score, and Area Under the Receiver Operating Characteristic Curve (ROC-AUC) were used. Accuracy measured overall classification correctness, while precision and recall assessed the model's ability to correctly identify AI-generated images. The F1-score provided a balance between precision and recall, and ROC-AUC measured the model's ability to distinguish between the two classes across different classification thresholds.

\subsection{Classification Performance}

Table~\ref{tab:classification_report} presents the classification report of the proposed model on the validation dataset.

\begin{table}
  \caption{Classification Report}
  \centering
  \begin{tabular}{lccc}
    \toprule
    Class & Precision & Recall & F1-score \\
    \midrule
    Human (0) & 1.00 & 0.87 & 0.93 \\
    AI (1) & 0.88 & 1.00 & 0.94 \\
    \bottomrule
  \end{tabular}
  \label{tab:classification_report}
\end{table}

The results indicated that the proposed model achieved strong performance across both classes. In particular, the model maintained a good balance between precision and recall, demonstrating its effectiveness in detecting AI-generated images while minimizing misclassifications of real images.

\subsection{Confusion Matrix Analysis}

\begin{figure}
  \centering
  \includegraphics[width=0.6\columnwidth]{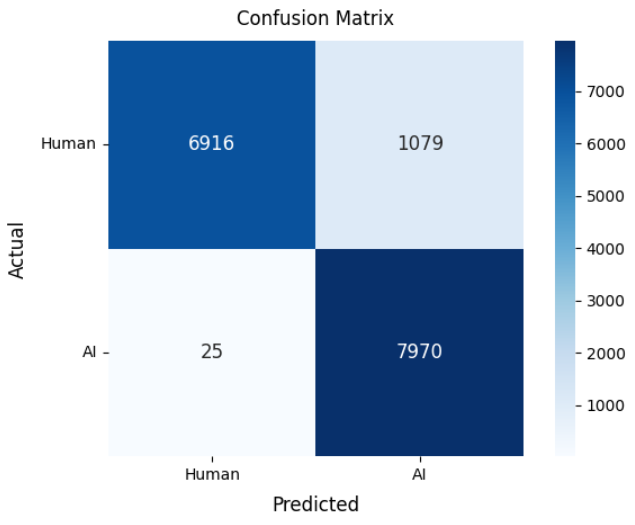}
  \caption{Confusion Matrix.}
  \label{fig:confusion_matrix}
\end{figure}

Figure~\ref{fig:confusion_matrix} shows the confusion matrix obtained on the validation dataset.

The confusion matrix provided insight into the classification performance of the model. A high number of true positives and true negatives indicated that the model correctly identified both AI-generated and real images. The relatively low number of false positives and false negatives suggested that the model effectively captured distinguishing features between the two classes.

\subsection{ROC Curve and AUC Score}

\begin{figure}
    \centering   
    \includegraphics[width=0.43\textwidth]{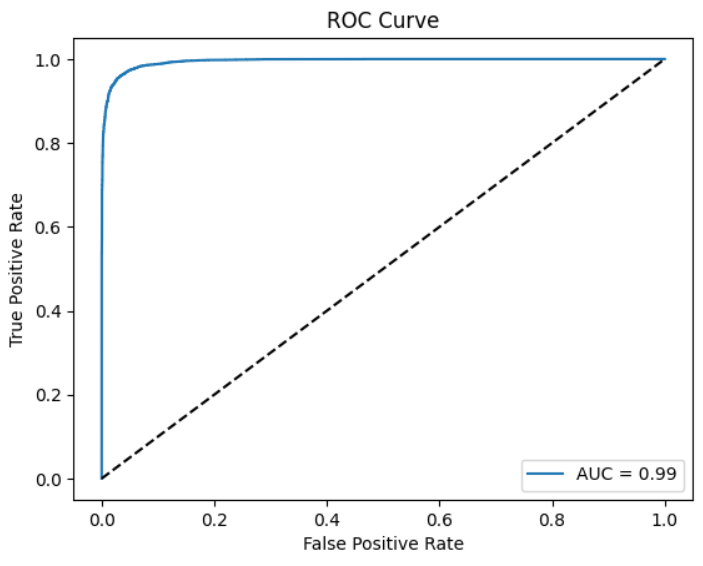}
    \caption{ROC Curve.}
    \label{fig:roc_curve}
\end{figure}

The Receiver Operating Characteristic (ROC) curve of the proposed model is illustrated in Figure~\ref{fig:roc_curve}.


The ROC curve illustrates the trade-off between the true positive rate and false positive rate across different thresholds. The high AUC score indicates that the proposed model effectively distinguishes between AI-generated and human-created images.

\subsection{Ablation Study}
\label{sec:ablation}

To validate the key design decisions of the proposed model, specifically the dual backbone architecture and the four channel RGB+Grayscale input representation, a systematic ablation study was conducted. Six model configurations were trained and evaluated under identical conditions (same hyperparameters, same dataset splits) to isolate the contribution of each component. Validation set metrics (accuracy, weighted precision, weighted recall, and weighted F1 score) were reported alongside the Kaggle competition score for the Detect AI vs. Human-Generated Images challenge ~\cite{kaggle_detect_ai}, which used F1 score as its primary evaluation metric. The competition score served as an independent measure of generalization to unseen test data. Table~\ref{tab:ablation} shows these results. Each experiment was repeated multiple times, and the reported values were the average to ensure statistical reliability.

\begin{table*}[t]
  \caption{Ablation Study: Component-Wise Performance Comparison}
  \centering
  \small
  \begin{tabular}{lccccc}
    \toprule
    Model Variant & Accuracy & Precision (Weighted) & Recall (Weighted) & F1 (Weighted) & Competition Score \\
    \midrule
    ConvNeXt Only (RGB) & 0.9876 & 0.9876 & 0.9876 & 0.9876 & 0.5526 \\
    ConvNeXt Only (RGB + Gray) & 0.9431 & 0.9473 & 0.9431 & 0.9430 & 0.7322 \\
    Swin Only (RGB) & 0.9551 & 0.9582 & 0.9551 & 0.9550 & 0.6632 \\
    Swin Only (RGB + Gray) & 0.9445 & 0.9493 & 0.9445 & 0.9443 & 0.7296 \\
    Combined Model (RGB Only) & 0.8787 & 0.9008 & 0.8787 & 0.8770 & 0.6878 \\
    Combined Model (RGB + Gray) & 0.9310 & 0.9386 & 0.9310 & 0.9307 & 0.7681 \\
    \bottomrule
  \end{tabular}
  \label{tab:ablation}
\end{table*}

Comparing each backbone variant with and without the grayscale channel revealed a consistent and significant improvement in the competition F1-score when the grayscale channel was added, despite a slight reduction in validation accuracy. This divergence between validation accuracy and competition score highlighted that high validation accuracy alone did not guarantee strong generalization to unseen data. The grayscale channel contributed luminance based structural and textural cues that were not fully captured by RGB channels, and these cues proved especially valuable for detecting the subtle spatial inconsistencies characteristic of AI generated images.

Comparing the full combined model against individual backbone variants using the same RGB+Grayscale input demonstrated the benefit of the dual backbone fusion strategy. While the single backbone variants (ConvNeXt + Gray: 0.7322, Swin + Gray: 0.7296) achieved higher validation accuracy than the full model, the full combined model achieved the highest competition score of 0.7681, surpassing both individual backbones. This result confirmed that ConvNeXt-Tiny and Swin-Tiny learned complementary representations and together produced a more generalizable model than either architecture alone. 

A notable observation across all variants was that validation accuracy did not reliably predict competition performance, suggesting severe overfitting to the validation distribution. This pattern underscored the importance of evaluating on an independent held out benchmark.

\section{Cross-Dataset Generalization Evaluation}
To evaluate the generalization capability of the proposed model beyond the training distribution, an external dataset, the AI vs Real Images Dataset~\cite{ai_real_images_dataset}, was used. It contained approximately 1000 images of real and AI-generated samples across five categories: nature, city, animals, food, and people. Real images were sourced from Unsplash, while AI images were generated using Stable Diffusion.

Unlike the training data, which consisted of paired real and synthetic images from a specific generation pipeline, this dataset introduced variations in both data sources and generation methods. Therefore, it provided a suitable benchmark for assessing model robustness under distribution shifts.

The model, trained solely on the original dataset without any fine-tuning, was evaluated on this external dataset. The overall performance is summarized in Table~\ref{tab:cross_dataset_results}, while the per-class performance is presented in Table~\ref{tab:per_class_results}.

\begin{table}[t]
  \caption{Cross-Dataset Evaluation Results}
  \centering
  \renewcommand{\arraystretch}{1.2}
  \begin{tabular}{lc}
    \toprule
    \textbf{Metric} & \textbf{Score} \\
    \midrule
    Accuracy & 0.8251 \\
    Precision (Weighted) & 0.8789 \\
    Recall (Weighted) & 0.8251 \\
    F1-score (Weighted) & 0.8353 \\
    \bottomrule
  \end{tabular}
  \label{tab:cross_dataset_results}
\end{table}

\begin{table}[t]
  \caption{Per-Class Performance}
  \centering
  \renewcommand{\arraystretch}{1.2}
  \begin{tabular}{lccc}
    \toprule
    \textbf{Class} & \textbf{Precision} & \textbf{Recall} & \textbf{F1-score} \\
    \midrule
    Human & 0.97 & 0.79 & 0.87 \\
    AI & 0.60 & 0.94 & 0.73 \\
    \bottomrule
  \end{tabular}
  \label{tab:per_class_results}
\end{table}

The results indicated an overall accuracy of 0.8251, with a weighted precision of 0.8789, recall of 0.8251, and F1-score of 0.8353.

The per-class analysis revealed an asymmetric performance pattern. For the Human class, the model achieved a precision of 0.97 and a recall of 0.79, resulting in an F1-score of 0.87. This indicated that predictions classified as real images were highly reliable; however, a notable proportion of real images were misclassified as AI-generated. In contrast, for the AI-generated class, the model achieved a high recall of 0.94 but a lower precision of 0.60, resulting in an F1-score of 0.73. This suggested that while the model was effective at identifying synthetic images, it tended to over-predict the AI class, leading to an increased number of false positives.

Overall, these findings suggested that the model learned features that partially generalized across datasets but remained sensitive to distributional differences between training and test data. The observed imbalance between precision and recall further indicated a bias toward detecting AI-generated content, likely due to variations in visual characteristics across datasets. This highlighted the importance of incorporating diverse generative models and data sources during training to improve robustness against unseen image synthesis techniques.

\section{Conclusion}

This paper presented a deep learning framework for detecting AI-generated images using an enhanced four-channel RGB+Grayscale representation and complementary feature extraction. The additional grayscale channel provided luminance and structural information alongside the original RGB features. ConvNeXt-Tiny and Swin-Tiny were employed in parallel to capture local spatial patterns and global contextual information, respectively. An attention-based fusion mechanism combined the representations learned by the two backbones before classification. 

Experimental results demonstrated that the proposed method achieved strong performance, with high accuracy, balanced precision and recall, and robust ROC-AUC characteristics. The model effectively distinguished between AI-generated and human-created images, even in the presence of subtle visual differences.

Overall, the findings highlighted the importance of combining multiple feature representations and advanced architectures to improve detection performance. As future work, the model can be extended to evaluate its generalization across different datasets and emerging generative models, particularly diffusion-based techniques. Additionally, incorporating frequency-domain features or self-supervised learning strategies may further enhance robustness and adaptability.

The proposed framework provided a reliable and scalable solution for AI-generated image detection, contributing to ongoing efforts in digital forensics and content authenticity verification.

\noindent\textbf{Code Availability: }https://github.com/Raleesa-Perera/Hybrid-ConvNeXt-Swin-model-for-AI-image-detection
\bibliographystyle{unsrt}  
\bibliography{references}

\end{document}